\newif\ifarxiv
\arxivtrue

\ifarxiv
  \documentclass[sigconf, screen, nonacm]{acmart}
\else
  \documentclass[sigconf, screen, review, anonymous]{acmart}
\fi
\AtBeginDocument{%
  }

\ifarxiv
\else
\setcopyright{none}
\copyrightyear{2026}
\acmYear{2026}
\acmDOI{}   
\acmConference[ICAIF '26]{7th ACM International Conference on AI in
  Finance}{November 14--17, 2026}{Milan, Italy}
\acmBooktitle{Proceedings of the 7th ACM International Conference on AI in
  Finance (ICAIF '26)}
\acmISBN{}
\fi

\usepackage[american]{babel}

\usepackage{mathtools}
\usepackage{graphicx}
\usepackage{booktabs}
\ifarxiv
  \usepackage[final, commandnameprefix=ifneeded]{changes}
\else
  \usepackage[commandnameprefix=ifneeded]{changes}
\fi

\begin{document}

\title{Amortized Interventional Forecasting for Multivariate CIR Processes}
\author{Andreas Sauter}
\authornote{Work partly carried out during an internship at ING.}
\correspondingauthor
\affiliation{%
  \institution{Vrije Universiteit Amsterdam}
  \city{Amsterdam}
  \country{Netherlands}}
\email{andreas@sauter.at}

\author{Sumit Sourabh}
\affiliation{%
  \institution{ING}
  \city{Amsterdam}
  \country{Netherlands}}

\author{Drona Kandhai}
\affiliation{%
  \institution{University of Amsterdam / ING}
  \city{Amsterdam}
  \country{Netherlands}
  }

\author{Erman Acar}
\affiliation{%
  \institution{University of Amsterdam}
  \city{Amsterdam}
  \country{Netherlands}}

\renewcommand{\shortauthors}{Sauter et al.}

\begin{abstract}
Mean-reverting dynamics are pervasive in finance, and the Cox--Ingersoll--Ross (CIR) process is a standard model for the time series they produce, from short rates to credit default swap (CDS) spreads. Yet CIR models capture only \emph{correlated} co-movement, not \emph{causal} influence between series, so they cannot answer the system's response when one series is externally shocked, which observational conditionals confound with historical co-movement. We make two contributions. First, an amortized model for distributional causal effect estimation that frames trajectories as time-stamped observations and predicts the calibrated multi-horizon shock response without retraining per scenario. Second, a causal multivariate CIR data-generating process that supplies the paired observational and interventional ground truth that real markets cannot. We instantiate and calibrate the framework on CDS spreads as a testbed. CIR-ACTIVA's validity is established on synthetic ground truth, independent of how well the simulator matches reality, while practical grounding is assessed by backtesting the generated traces against real CDS data. Against observational and amortized causal-inference baselines, CIR-ACTIVA leads on both causal selectivity in the joint distribution and horizon-resolved calibration, retaining its selectivity once the interventional law varies over the horizon, with gains concentrating at short horizons. This opens up a class of what-if queries on coupled spread systems, CDS stress testing among them, that observational forecasters cannot answer.
\end{abstract}



%

\maketitle

\section{Introduction}
Many financial time series like short rates, default intensities, and stochastic volatility are strictly positive and mean-reverting, and the CIR model is their standard modelling tool~\citep{Brigo2006InterestPractice,Cox1985ARates,Heston1993}. Its multivariate extensions reproduce co-movement across series~\citep{ChenScott1993Multifactor,DuffieGarleanu2001CDO}, but only through correlation, with no notion of one series \emph{causally} influencing another. Many decisions about such systems are nonetheless causal: they ask how the rest of the system responds when one series is externally shocked, a causal question that correlation among historical paths cannot answer.

We study this interventional problem for multivariate CIR processes. The target is an interventional distribution, the response of the remaining series when one series' dynamics are externally modulated, rather than a conditional forecast of future values among historical paths in which that series happened to take a given level. The two coincide only under restrictive assumptions, because a high value may be a symptom of shared stress rather than its cause. We instantiate the problem on credit default swap (CDS) spreads~\citep{Hull2000ValuingRisk,Giglio2012CreditRisk}, where spread forecasting is well studied but framed non-causally~\citep{Vukovic2022AreAutoregression,Mao2023ForecastingLSTMs,Liu2022AMechanism,Kubiak2025MacroVAE:Conditioning}. There, the intervention is a stress test that forces one name's spread to a stressed level and asks how the others respond. The distinction is operational: a model that answers the conditional query instead assigns a response to names the shock cannot causally reach, so a stress test built on it books losses and sizes hedges against exposures that do not exist.

Recent causal foundation models create a new opportunity for this problem by providing amortized estimators of interventional distributions~\citep{Sauter2025ACTIVA:Autoencoder,Balazadeh2025CausalPFN:Learning,Dhir2026EstimatingMeta-Learning}. At the same time, recent tabular foundation-model approaches suggest that time series can be represented as time-stamped tabular prediction problems~\cite{Hoo2025TheFeatures}, bridging time-series forecasting and tabular causal inference. We build on this perspective to develop CIR-ACTIVA, a causal forecasting model for multivariate CIR shock scenarios. 
Our model consumes only CIR-type trajectories, so it is not specific to CDS; credit spreads are simply the mean-reverting series we calibrate and evaluate. Evaluation is the central obstacle: ground-truth interventional distributions are unobserved in real CDS data. We therefore introduce a causal multivariate CIR data-generating process that preserves key properties of credit-spread series while allowing directed causal links between components, supplying the paired observational and interventional samples that are needed for amortized causal inference.

This paper makes two contributions:  1) \textbf{CIR-ACTIVA model (Section~\ref{sec:cir_activa}).} We develop a model for amortized distributional causal effect estimation that treats spread trajectories as time-stamped observations and uses a horizon-bucketed decoder, while leveraging  implicit model averaging. 2) \textbf{Causal CIR simulator (Section~\ref{sec:data_generation}).} To train and evaluate it, we introduce a causal multivariate CIR data-generating process that produces paired observational and interventional samples, providing a benchmark for interventional forecasting. 
 Empirically (Section~\ref{sec:results}), CIR-ACTIVA leads every baseline on both causal selectivity in the joint distribution, shielding non-affected series from phantom shock effects, and horizon-resolved calibration, with gains concentrated at the short horizons which are most relevant for stress testing.
\ifarxiv
Code for the model, the causal CIR simulator, and the experiments is
available at \url{https://github.com/sa-and/cir-activa}.
\fi

\section{Background and Notation}\label{sec:background}
In this section, we introduce the theoretical foundations of our approach along with the notation. We first review structural causal models and soft interventions, then describe amortized causal effect estimation, and summarize the CIR process and its role in financial time-series modeling.

\subsection{Causal Models}\label{sec:causal_models}
A \emph{Structural Causal Model} (SCM)~\citep{Pearl2009Causality} over endogenous variables $\boldsymbol{V}=\{V_1,\ldots,V_d\}$, and exogenous variables $\boldsymbol{U} =\{U_1, \ldots, U_m\}$ assigns to each variable $V_i$ a \emph{structural equation} $V_i \leftarrow f_i(Pa_{V_i}, Exo_{V_i})$, expressing it as a deterministic function of its \emph{causal parents} $Pa_{V_i}$ (the variables that directly influence $V_i$) and independent exogenous noise terms  $Exo_{V_i} \subseteq \boldsymbol{U}$, with $U_i \perp\!\!\!\perp U_j$ for $i \neq j$. Running the SCM without external manipulation yields \emph{observational data}. We  employ \emph{soft} interventions~\citep{Eberhardt2007Interventions}, which replace the mechanism $f_i$ of the target with a modified mechanism. Concretely, $do\!\left(V_i{:=}(1+\delta)f_i\right)$ rescales the target's own dynamics by a magnitude $\delta$ while \emph{retaining} its dependence on $Pa_{V_i}$, so the intervened variable still responds to its causes. We write $do(V_i)$ for this soft intervention throughout, with the magnitude $\delta$ carried by the representation $\boldsymbol{i}$.

Given observational data $\boldsymbol{D}^{\mathcal{M}_o}$ sampled from the observational distribution $p_{\mathcal{M}_o}(\boldsymbol{V})$ of the (unknown) generating model $\mathcal{M}_o$ and an intervention query $do(V_i)$, causal effect estimation seeks the post-interventional distribution $p(\boldsymbol{V} \mid \boldsymbol{D}^{\mathcal{M}_o}, do(V_i))$. A fundamental complication is \emph{observational equivalence}: many causal models can generate the same observational distribution while implying different interventional distributions.  Rather than committing to a single model, a principled approach is to represent this ambiguity as a mixture over all observationally equivalent models~\cite{Sauter2025ACTIVA:Autoencoder}:
\begin{equation}\label{eq:mixture}
  p\!\left(\boldsymbol{V} \mid \boldsymbol{D}^{\mathcal{M}_o}, do(V_i)\right)
  = \int p_\mathcal{M}\!\left(\boldsymbol{V} \mid do(V_i)\right) p\!\left(\mathcal{M} \mid \boldsymbol{D}^{\mathcal{M}_o}\right) d\mathcal{M}
\end{equation}

\subsection{Amortized Causal Effect Estimation}

Classical estimators solve causal effect estimation from scratch for every dataset. \emph{Amortized} approaches instead cast it as a distribution over \emph{tasks}: a simulator draws a causal model $\mathcal{M}\sim p_{\mathrm{tr}}(\mathcal{M})$ from a \emph{training distribution} over SCMs and emits a task consisting of an observational context, an intervention query, and paired interventional data. A single model is trained across many such tasks to map context and query to the interventional distribution, with the interventional samples supervising training but withheld at inference. Test-time estimation on a new task is then a single forward pass with no per-dataset re-training, valid under the assumption that test tasks are drawn from the same $p_{\mathrm{tr}}(\mathcal{M})$. Because the simulator can produce many observationally equivalent SCMs with differing interventional behaviour, this regime also biases the model toward the model-averaging mixture in Eq.~\eqref{eq:mixture}, representing causal ambiguity in its predictions rather than collapsing to a point estimate. ACTIVA~\citep{Sauter2025ACTIVA:Autoencoder} and MACE-TNP~\citep{Dhir2026EstimatingMeta-Learning} instantiate this paradigm.

\subsection{CIR Model in Finance}
The CIR model~\cite{Cox1985ARates} is a non-negative, mean-reverting stochastic process as $
  dx_t = \kappa(b - x_t)\,dt + \sigma\sqrt{x_t}\,dW_t,$
where $\kappa > 0$ is the mean-reversion speed, $b > 0$ the long-run mean, $\sigma > 0$ the volatility, and $W_t$ a standard Brownian motion. The CIR process is used across finance for short rates, default intensities, and stochastic volatility~\citep{Cox1985ARates,DuffieGarleanu2001CDO,Heston1993}, and is a standard tool for modeling CDS spreads~\citep{Brigo2006InterestPractice}.

Multivariate extensions stack several CIR factors so that each spread $x^{(i)}_t$ is driven by a combination of shared and idiosyncratic square-root factors~\citep{ChenScott1993Multifactor,DuffieGarleanu2001CDO}. In the canonical formulation each spread follows its own mean-reverting equation, and co-movement between series arises through the common factors. This reproduces empirical co-movement in spread panels but only through correlation, with no representation of causal influence between spreads.

\section{Causal CIR model}\label{sec:causal_cir}

Entities in a multivariate CIR model do not merely co-move; they can interact \emph{causally} through concrete financial mechanisms. A prominent example is \emph{sector contagion}: an idiosyncratic credit event at one automotive manufacturer propagates to its sector peers, raising their spreads even absent any common macro factor~\citep{Jorion2007GoodContagion}. Ignoring this causal structure can lead to mispriced hedges and spurious intervention predictions.

We formalize these interactions by extending the single-series CIR model of Section~\ref{sec:background} to a \emph{multivariate causal CIR model} defined as an SCM over $d$ variables. Each variable $i$ evolves according to a structural equation analogous to the CIR dynamics:
\begin{equation}
  V_{i}^{t+1} = V_{i}^t
    + \left(1-e^{-\kappa_{i}}\right)\!\left(b_{i}^{t+1} - V_{i}^t\right)
    + s_{i}^t\;\varepsilon_{i}^t,   
\end{equation}
where $\varepsilon_i^t$ is a noise term based on exogenous variables $\{U_j^t\}_{j}\in Exo_{V_i}$ with $U_j^t \overset{\mathrm{iid}}{\sim} \mathcal{N}(0,1)$, rescaled to unit variance. Both coefficients come from the one-step solution of the continuous-time dynamics: the drift closes the exact fraction $1-e^{-\kappa_i}$ of the gap to $b_i^{t+1}$, and $(s_i^t)^2 = \sigma_i^2 V_i^t (1-e^{-2\kappa_i})/(2\kappa_i)$ integrates the diffusion over the step with its state dependence held at $V_i^t$, coinciding with the exact CIR conditional variance when $V_i^t = b_i^{t+1}$. 

The key structural distinction from a multivariate CIR model is the \emph{causal long-run mean} $b_i^{t+1}$, which is not a fixed scalar but combines the contemporaneous values of variable $i$'s causal parents $Pa_{V_i}$ with an explicit time modulation $c_i(t)$:
\begin{equation}
  b_i^{t+1}
  = c_i(t) \left( \sum_{j\,\in\,Pa_{V_i}} w_{i,j}\, V_j^{t+1}
    + w_{i,0}\,\bar{b}_i\right),
\end{equation}
where weights $\boldsymbol{w}_{i} = \mathrm{softmax}(\boldsymbol{s}_{i})$ with $\boldsymbol{s}_{i} \sim \mathrm{Uniform}(0.3, 3)^{|Pa_{V_i}|+1}$ are normalized so that the target mean remains anchored near the baseline $\bar{b}_i$ while allowing causal parents to shift the equilibrium level, and $c_i(t) > 0$ is a per-variable \emph{time-modulation factor} applied multiplicatively to the long-run target. 
Parents enter contemporaneously, so a directed cycle forms a simultaneous system within a step. When $Pa_{V_i} = \emptyset$ and $c_i(t) = 1$, the model reduces to a standard univariate CIR process with $b_i^{t+1} = \bar{b}_i$.


\section{Data generation}\label{sec:data_generation}
In this section we describe how we construct the simulator distribution $p_{\mathrm{tr}}(\mathcal{M})$ over CDS spreads via our causal CIR model and evaluate its match to real data. 

\subsection{Calibration and Sampling of CIR Parameters}\label{sec:cir_calibration}
To obtain realistic parameters for our data generator, we calibrate on a corpus of $661$ single-name 5Y CDS series (daily, 2015--2021), split into quarter-length chunks.
We observe each CDS series at unit time intervals and calibrate its CIR parameters $(\kappa,b,\sigma)$ by maximum likelihood from the closed-form noncentral chi-square one-step transition density~\citep{overbeck97,Cox1985ARates}. 





After series-wise MLE, we model a distribution over the estimated parameters and the observed initial spread level $x^0$ by a Gaussian mixture for $\boldsymbol{\psi}=\big(\log\kappa,\log b,\log\sigma,\log x^0\big)$,
estimated by Expectation–Maximization~\citep{Dempster1977MaximumAlgorithm}. The log-space fit enforces positivity, the mixture captures multi-modality, and jointly fitting $x^0$ starts trajectories from realistic levels. For generation, CIR parameters for each variable  are drawn independently from the fitted Gaussian $p(\boldsymbol{\psi})$. 
Because the calibrated $\bar{b}_i$ is no longer a self-contained long-run mean in the causal model, some parameter draws produce non-reverting, diverging trajectories. To exclude these from the training set, any simulation where at least one spread leaves the interval $[-0.1,\,10]$ or $\bar{b}_i > 5$ is discarded. 

\subsection{Data Generation Procedure}
We generate each SCM, using the CausalPlayground library~\citep{Sauter2024CausalPlayground:Research}, over $d = 3$ spread variables with  probability of each causal relation of $50\%$; directed cycles are permitted and resolved by unrolling the cycle into a fixed number of acyclic layers. Confounding is enabled: each exogenous noise variable can affect multiple endogenous variables, inducing correlated residuals across variables without an explicit causal arc.
A dataset consists of an observational run of $T_{\mathrm{obs}} = 200$ steps recorded after a warm-up of $100$ discarded steps. At the end of the observational run, a single variable $i$ is chosen uniformly at random as the intervention target. We apply a \emph{soft} intervention as defined in Section~\ref{sec:causal_models} 
with $\delta \sim \mathcal{N}(0.20,\, 0.05^2)$  a multiplicative shift drawn once per dataset and held fixed, mirroring how credit stress scenarios are posed in practice
~\citep{Glasserman2015StressScenario}. From the last observational state, $K_{\mathrm{int}} = 5$ independent interventional trajectories of length $T_{\mathrm{int}} = 100$ steps are then simulated under this intervention as samples of the interventional distribution. 

For per-instance temporal variation and efficiency, we save only a uniformly random subsample of $S_{\mathrm{obs}} = 50$ observational and $S_{\mathrm{int}} = 30$ interventional steps per trajectory.
The resulting triple is ($\boldsymbol{D}^\mathcal{M}$, $\boldsymbol{D}^{\mathcal{M}_{do(V_i)}}$, $\boldsymbol{i}$), where $\boldsymbol{D}^\mathcal{M} = \{\boldsymbol{v}^\mathcal{M}_n\}_{n=1}^{S_{\mathrm{obs}}} \sim  p_\mathcal{M}(\boldsymbol{V})$ are the observational samples, $\boldsymbol{D}^{\mathcal{M}_{do(V_i)}} = \{\boldsymbol{v}^{\mathcal{M}_{do(V_i)}}_n\}_{n=1}^{5 \times S_{\mathrm{int}}} \sim p_\mathcal{M}(\boldsymbol{V} \mid do(V_i))$ the interventional samples, and $\boldsymbol{i}\in\mathbb{R}^d$ holds $\delta$ at the intervened dimension and $0$ elsewhere. This triple constitutes a single amortization \emph{task}, induced by an SCM $\mathcal{M}\sim p_{\mathrm{tr}}(\mathcal{M})$. 

Five datasets are drawn from each distinct SCM before a new SCM is sampled, yielding structural diversity. The full corpus of 5000 datasets is split 70\,/\,15\,/\,15 into training, validation, and test sets. The split is over instances, not SCM identity, so a given SCM's five datasets may fall across multiple splits; this matches the same-distribution generalization claimed above.



\subsection{Time-Modulation Variants}\label{sec:time_modulation}
We study two instantiations of the time-modulation factor $c_i(t)$. In the \emph{non-cyclical} variant the modulation is switched off, $c_i(t) = 1$, so the long-run target is driven purely by the causal parents and is stationary in time. In the \emph{cyclical} variant we use a sinusoidal modulation
$
  c_i(t) = 1 + A_i \sin(\omega_i t + \phi_i),
$
where the amplitude $A_i \sim \mathrm{Uniform}(0.2,\, 0.5)$, the frequency $\omega_i = 2\pi n_{\mathrm{cyc}}/T_{\mathrm{int}}$ with $n_{\mathrm{cyc}} \sim \mathrm{Uniform}(0,\, 3)$ full cycles over the interventional window $T_{\mathrm{int}}$, and the phase $\phi_i \sim \mathrm{Uniform}(0,\, 2\pi)$ are all sampled independently per variable. The modulation is mean-preserving across the task distribution: the uniform phase gives $\mathbb{E}[c_i(t)] = 1$ at every $t$ and for every $\omega_i$, so the marginal long-run level remains anchored.
The cyclical variant stress-tests whether a model can track a changing mean across the interventional horizon, a capability that goes beyond capturing time-varying variance alone.


\subsection{Distance to the Real  Market Data}\label{sec:backtest}

We assess how far the simulated traces reproduce real CDS data, following standard validation for synthetic financial data~\citep{Cont2001Empirical, Assefa2020Generating}. The reference is our calibration corpus. We subsample the reference on the same time stamps as our generated data. 
We report the lag-1 autocorrelation of spread levels, expressed as the implied half-life of a level shock, and the frequency of large log-returns. The comparison is necessarily on the per-name marginals: real data does not reveal which names are causally linked, so the simulator's dependence structure has no multivariate counterpart.

We find that generated levels lie in the empirical range (median $49$--$52$ against $64$\,bp), and a shock decays with a half-life of $8.9$ days in the cyclical variant and $2.5$ days without mean modulation, against $21$ days in the panel, while a $3\sigma$ log-return appears in $36\%$ of generated windows against $77\%$ of real ones. Thinner tails are expected of a pure diffusion, as reproducing the fat tails seen in real
spreads requires a jump component~\citep{Merton1976Option}, a documented driver of CDS
spreads~\citep{Zhang2009Explaining}. The faster mean reversion is instead an artifact of the parameter prior: the autoregressive coefficient is downward-biased on the quarter-length windows we fit~\citep{Kendall1954Note}, so the prior is centred on faster reversion than the panel exhibits. Absent an interventional market counterpart, post-intervention traces are compared with
real windows of matched length. Both variants behave as they do observationally, the
cyclical one closer to the panel and the non-cyclical one reverting faster and with fewer
large moves, so intervening introduces no discrepancy beyond the two already noted. In both
variants the response shifts the target's level by a median of $+62\%$ without mean modulation and $+70\%$ with it, exceeding
$90\%$ of 100-day moves in the panel.

Importantly, neither of the two gaps described above concern our estimator, which is amortized over whatever simulator provides its training tasks, so closing them means retraining, not redesign. Both gaps make the task easier than reality, since shorter memory reduces confounding history, and thinner tails reduce extreme moves, so the model evaluation best read as an optimistic bound on the real data.

\section{CIR-ACTIVA Model Description}\label{sec:cir_activa}

This section describes our approach to amortized interventional distribution estimation for multivariate CIR time series. We adopt ACTIVA~\citep{Sauter2025ACTIVA:Autoencoder} as our generative backbone and keep its amortized estimator unchanged. ACTIVA's amortized inference over the training distribution $p_\mathrm{tr}(\mathcal{M})$ and its implicit model averaging over observationally equivalent SCMs are properties of that estimator and its training distribution, not of any particular architecture or output likelihood; our modifications change only the latter two, so both properties carry over unchanged.

Our two main modifications are a \emph{time-series framing} that promotes time to a variable of the model and a \emph{horizon-bucketed decoder} whose predicted interventional law varies with forecast distance. Together with the inherited guarantees above, these form the core of CIR-ACTIVA. Three supporting choices (a conditional normalizing-flow prior, per-variable multi-channel latent codes, and a $\sqrt{\cdot}$ output transform for non-negative spreads) round out the model.

\subsection{Base model (ACTIVA)}\label{sec:activa}
Following ACTIVA~\citep{Sauter2025ACTIVA:Autoencoder}, we model amortized causal effect estimation as a $\beta$-CVAE with generative process
\begin{equation}
  p_\theta(\boldsymbol{V}, \boldsymbol{z} \mid \boldsymbol{D}^\mathcal{M}, \boldsymbol{i}) = p_\gamma(\boldsymbol{V} \mid \boldsymbol{z})\, p_\eta(\boldsymbol{z} \mid \boldsymbol{D}^\mathcal{M}, \boldsymbol{i}),
\end{equation}
conditioning on an observational context $\boldsymbol{D}^\mathcal{M}$ and an intervention $do(V_i)$, represented by $\boldsymbol{i}$, the model produces a latent code $\boldsymbol{z}$ via the learned prior $p_\eta$ trained jointly with a variational posterior $q_\phi$. Both are parameterized by transformer encoders: the prior encoder processes only the observational data and intervention query; the posterior encoder additionally processes the interventional samples. A transformer decoder maps sampled $\boldsymbol{z}$ to a GMM over the interventional distribution.

\subsection{Time-Series Framing}
Existing methods treat an observational context as an unordered set of i.i.d.\ samples from a single SCM, with no notion of time~\cite{Sauter2025ACTIVA:Autoencoder,Dhir2026EstimatingMeta-Learning}. We keep this exchangeable sample view and accommodate time series through a minimal change of perspective: time is promoted to a variable of the model. Concretely, we let $\boldsymbol{V} = (\boldsymbol{Y}, T)$, where $\boldsymbol{Y}$ collects the $d$ series variables and $T$ is the time stamp. The observational context $\boldsymbol{D}^\mathcal{M}$ has exactly these columns; each row is therefore a single timestamped snapshot $(\boldsymbol{y}, t)$. Because every row carries its own timestamp, the rows remain exchangeable: permuting them leaves the model's prediction unchanged. Conditional on the latent $\boldsymbol{z}$ and the timestamp, the rows are then independent up to a within-trajectory residual that decays with the sampling gap.

The framing is information-preserving: timestamps make each observation's temporal position explicit, so no temporal position information is lost by presenting the observations as an unordered set. It mirrors a line of recent work that casts forecasting as tabular prediction with time as an input feature, most directly TabPFN-TS~\citep{Hoo2025TheFeatures}. This framing lets us draw on causal-inference results not previously accessible for time-series problems. The trade-off is that temporal structure is no longer imposed by the architecture but must be \emph{learned}. The full burden of discovering how the spread variables depend on $T$ 
falls on the model itself. 

\subsection{Encoding Architecture}\label{sec:encoder_and_posterior}
Each input row carries variable values, its time index, and an intervention channel; in every projection the value and time dimensions pass through separate MLPs, so the model treats temporal position and spread magnitude as distinct signals. Two weight-identical equivariant transformer encoders~\citep{Lorch2022AmortizedLearning,Sauter2025ACTIVA:Autoencoder} embed the two input paths (the observational context $\boldsymbol{D}^\mathcal{M}$ and, during training only, the interventional samples $\boldsymbol{D}^{\mathcal{M}_{do(V_i)}}$) both receiving the masked intervention values $\boldsymbol{i}$, and mean-pool over the sample axis to give per-variable embeddings that are invariant to sample order and equivariant under variable permutations. Concatenating the two embeddings and passing them through a conditioning transformer yields the posterior $q_\phi(\boldsymbol{z} \mid \boldsymbol{D}^{\mathcal{M}_{do(V_i)}}, \boldsymbol{D}^\mathcal{M}, \boldsymbol{i})$, a diagonal Gaussian over a per-variable latent $\boldsymbol{z} \in \mathbb{R}^{d \times L}$ with $L$ channels per variable, enough for the decoder to express a per-variable time-shape rather than only a level. The prior $p_\eta(\boldsymbol{z} \mid \boldsymbol{D}^\mathcal{M}, \boldsymbol{i})$ consumes the same observational embedding, and hence the same intervention query, and is a conditional continuous normalizing flow~\citep{Chen2018NeuralODEs,Grathwohl2019FFJORD}: it transports a standard normal through a velocity field parameterized by the same equivariant transformer, integrated with a fixed-step Runge--Kutta scheme, with log-density from the instantaneous change-of-variables formula. The flow lets the prior represent a multimodal $p(\boldsymbol{z} \mid \boldsymbol{D}^\mathcal{M}, \boldsymbol{i})$ without a fixed number of components while preserving variable-permutation equivariance; the ablation replaces it with a conditional diagonal Gaussian-mixture prior.

\subsection{Decoder}
Our transformer decoder maps the sampled latent $\boldsymbol{z}$ to the parameters of a joint distribution $p_\gamma(\boldsymbol{V} \mid \boldsymbol{z})$ over the full $(d + 1)$-dimensional output $\boldsymbol{V} = (\boldsymbol{Y}, T)$ of CIR variables and time. 
The decoder partitions the forecast horizon $[t_{\min}, t_{\max}]$ into $K$ equal-width time buckets and emits, per bucket $k$, a full-covariance Gaussian over the spread variables together with a linear-in-time mean shift, which serves as the data-conditional factor of the joint:
\begin{equation}
  p_\gamma(\boldsymbol{Y} \mid t, \boldsymbol{z}) = \mathcal{N}\big(\boldsymbol{Y};\ \mu_k(\boldsymbol{z}) + A_k(\boldsymbol{z})\,(t - \tau_k),\ \Sigma_k(\boldsymbol{z})\big),
\end{equation}
where $k = \lfloor (t - t_{\min}) / \Delta \rfloor$ indexes the bucket of width $\Delta$ centered at $\tau_k$; the mean $\mu_k(\boldsymbol{z}) \in \mathbb{R}^d$, the time-slope $A_k(\boldsymbol{z}) \in \mathbb{R}^d$, and the full covariance $\Sigma_k(\boldsymbol{z}) \in \mathbb{R}^{d \times d}$ are emitted by the decoder and are equivariant under variable permutations. With the time marginal $p(t)$ fixed uniform on $[t_{\min}, t_{\max}]$, the displayed Gaussian is the conditional factor of a \emph{joint} density $p_\gamma(\boldsymbol{V} \mid \boldsymbol{z}) = p_\gamma(\boldsymbol{Y} \mid t, \boldsymbol{z})\,p(t)$ that the model maximizes over $(\boldsymbol{Y}, T)$, not over the conditional alone. Fixing $p(t)$ rather than learning it is a deliberate design choice to provide an inductive bias about known dynamics of time. 
This horizon bucketing is the decoder's core temporal adaptation: it represents how the interventional law changes with forecast distance, and conditioning on any query horizon stays closed-form. 

The only CIR-specific component is the output transform: the Gaussians are placed over the \emph{square root} of the spreads (time is not transformed) and mapped back through the change of variables, with all likelihoods reported in raw units. For CIR-type diffusions the $\sqrt{\cdot}$ variance-stabilizes the conditional noise while squaring back enforces positivity.


\subsection{Training and Hyperparameters}
The model is trained by maximizing the $\beta$-CVAE objective, the KL-weighted ELBO:
\begin{equation}
  \mathcal{L} = \mathbb{E}_{q_\phi}\!\left[\log p_\gamma(\boldsymbol{v}^{\mathcal{M}_{do(V_i)}} \mid \boldsymbol{z})\right] - \beta\,\mathrm{KL}\!\left(q_\phi \,\|\, p_\eta\right),
\end{equation}
with the KL normalized by the number of time steps and given a deliberately large weight $\beta$. The motivation is the inference-time path: an interventional query draws $\boldsymbol{z}$ from the conditional prior, never the posterior, so predictive information hidden in trajectory-specific latent codes is unavailable exactly when needed; a large $\beta$ makes such storage costly and forces predictive structure into the conditioning path of observational context and intervention. 
To prevent latent collapse we floor the KL at a small free-bits budget per sample and ramp $\beta$ over an initial warmup phase. 
For model selection we retain the checkpoint with the lowest prior-path predictive NLL on the held-out validation set.

For our network architecture, we use a hidden width of $128$ throughout, per-variable embedding dimension $28$, $4$ attention heads, $4$ encoder and $4$ decoder layers, dropout $0.1$, $L = 8$ latent channels per variable, $K = 8$ time buckets (decoder mixture components), a $2$-layer flow velocity field integrated with $8$ Runge--Kutta steps for the prior, KL weight $\beta = 25$ with a $0.5$-nat free-bits floor, and AdamW (learning rate $10^{-3}$, cosine decay\ with linear warmup, global gradient clipping at norm $1.0$) for $5{,}000$ epochs at batch size $128$ with a sigmoid $\beta$-warmup over the first $4{,}500$ epochs. The latent-scale guardrails are a smooth ceiling of $3.0$ on the posterior standard deviation and of $5.0$ on the flow-field velocity magnitude, and all computation is carried out in \texttt{bfloat16} precision.

\section{Experimental Setup}

\paragraph{Datasets and Baselines.}
We evaluate on the held-out test splits of the two causal CIR datasets of Section~\ref{sec:data_generation}: the \emph{non-cyclical} variant and the \emph{cyclical} variant.
We compare our CIR-ACTIVA against four baselines: the original \textbf{ACTIVA} model, a multi-variable variant of MACE-TNP, a Gaussian Mixture Regression (GMR) baseline, and an oracle performance floor.
\textbf{MACE-TNP} is the transformer neural-process estimator introduced by~\cite{Dhir2026EstimatingMeta-Learning}, used here in its joint mixture-of-Gaussians variant~\cite{Sauter2025ACTIVA:Autoencoder}. Like our model, it amortizes across SCMs and uses intervention-magnitude conditioning, but it lacks any time-series-specific structure. We are not aware of a prior application of it to interventional forecasting of a
temporal process, so we apply it here ourselves. Sharing our amortized causal framework and
differing mainly in temporal structure, it is the most informative comparison in our study. The \textbf{GMR} baseline is non-causal: for each instance it fits a full-covariance Gaussian mixture ($K = 8$ components) to the observational context, then answers an interventional query by \emph{conditioning} every component analytically on the intervened variable taking its intervention value and samples from the resulting conditional mixture. Because it conditions rather than intervenes, it propagates observational associations to every correlated variable regardless of causal role; it is included as a strong non-causal reference, not as a valid interventional estimator.
The \textbf{Oracle} reference is a leave-one-out empirical floor: each ground-truth point is scored against the remaining points of the same instance, so the scoring ensemble and its target are draws from one law, where every proper score is at its expected minimum. The floor is strict because that law is the instance's \emph{known} SCM, which the context identifies only partially, so no model conditioned on context and intervention alone attains it. For the conditional scores the ensemble is restricted to the same $t$. Because the floor is subtracted from every method within a bin, it shifts all curves equally and leaves method-to-method differences and ranking unchanged.

\paragraph{Metrics}\label{sec:metrics}
Each metric probes a different aspect of the predictive distribution; all  scores are sample-based, i.e. we compare samples from the predicted distribution with the ground-truth samples; and lower is better throughout.

We evaluate the \emph{joint} over variables. Furthermore, we score the exact-time conditional $p(\boldsymbol{Y} \mid t)$ by conditioning the predicted joint on each recorded $t$ analytically, aggregating into 5 equal-width horizon bins. This isolates \emph{temporal calibration}, complementary to the joint scores, which instead test how mass is allocated \emph{across} horizons. Every joint metric is additionally reported for all variables (\emph{full}), causal descendants of the intervened variable (\emph{desc}), and non-descendants (\emph{nondesc}), testing whether effects reach descendants while sparing the rest.

The Energy score~\citep{Gneiting2007StrictlyEstimation} summarizes distributional spread and bias, the Variogram score~\citep{Scheuerer2015VariogramBasedQuantities} is sensitive to the cross-variable/cross-time dependence structure, and the CRPS~\citep{Gneiting2007StrictlyEstimation} is the familiar per-variable marginal score that the two multivariate scores generalize (being univariate, it carries no dependence information, so we report it pooled over variables rather than decomposed by causal role). Energy and Variogram are standardized per dimension, and the scores are calculated comparing samples from the predicted distribution to the respective ground-truth samples. All reported intervals are marginal $95\%$ confidence intervals over test instances.

\section{Results}\label{sec:results}
We assess each method on two axes that an interventional model must satisfy simultaneously: the \emph{whole-horizon joint} (does it reproduce the correct joint over variables and time, respecting causal structure?) and \emph{temporal calibration} (resolved by horizon, does it match the per-horizon conditional $p(\boldsymbol{Y}\mid\mathrm{horizon})$?).

\begin{figure}
    \centering
    \includegraphics[width=\linewidth]{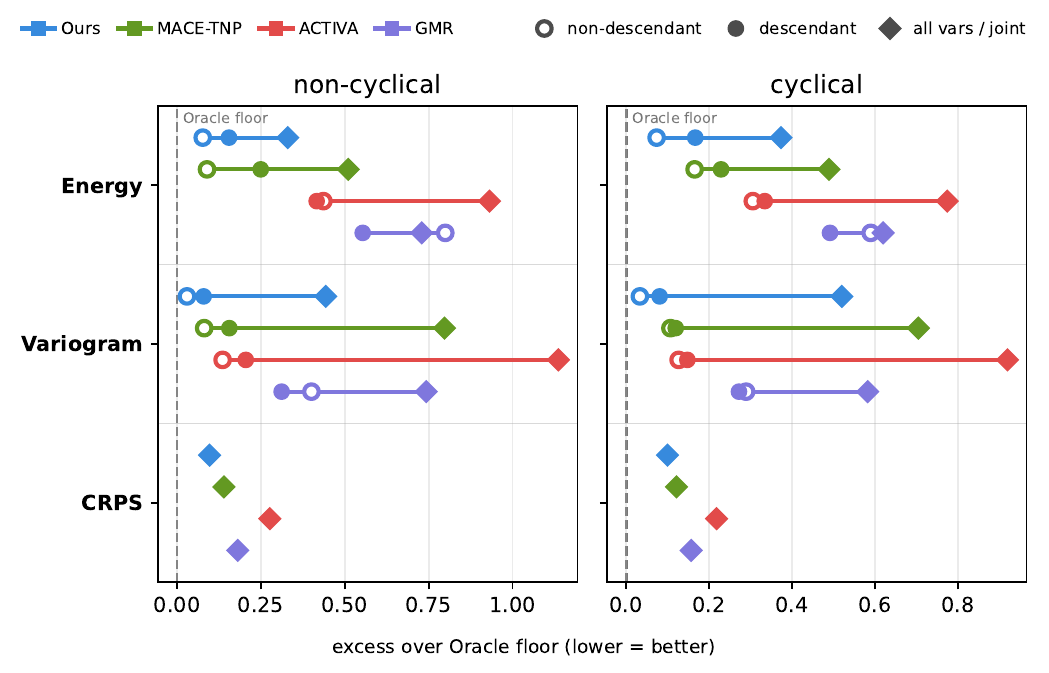}
    \caption[]{Whole-horizon scores as \emph{excess over the Oracle floor} (dashed line at $0$) for the non-cyclical (left) and cyclical-mean (right) CIR datasets, in Energy, Variogram and CRPS blocks. In the Energy and Variogram blocks each method's connected markers trace non-descendants (open circle), descendants (filled circle) and all variables (\emph{joint}, diamond); CRPS (time excluded) is a single per-method marginal.}
    \Description{Dumbbell plot with two panels, non-cyclical on the left and
    cyclical on the right, sharing a horizontal axis of score excess over the
    Oracle floor where zero is best. Three metric blocks are stacked
    vertically: Energy, Variogram and CRPS. Within the Energy and Variogram
    blocks each of the four methods occupies one row, drawn as a line joining
    an open circle for non-descendants, a filled circle for descendants and a
    diamond for the joint score. Our method sits closest to zero in every
    block and in both panels, followed by MACE-TNP, then ACTIVA, with GMR
    intermediate. For our method, MACE-TNP and ACTIVA the open non-descendant
    marker lies to the left of the filled descendant marker, so unaffected
    variables are scored better than affected ones. For GMR the order is
    reversed in both panels: its non-descendant marker lies to the right of
    its descendant marker, indicating that it assigns shock responses to
    variables the intervention cannot reach. The CRPS block shows a single
    diamond per method, ordered the same way with ours nearest zero.}
    \label{fig:joint_split_scores}
\end{figure}

\subsection{Whole-Horizon Effect Propagation}\label{sec:joint_scores}

Figure~\ref{fig:joint_split_scores} reports the whole-horizon joint scores: the Energy and Variogram Scores decomposed by the causal role of each variable relative to the intervention target alongside the per-variable CRPS (time excluded) as a marginal summary. A faithful interventional model should propagate the effect to descendants while leaving non-descendants close to their observational behaviour. 

On the pooled joint our method leads MACE-TNP on both datasets: Energy $0.33$ against $0.51$ and Variogram $0.44$ against $0.80$ on the non-cyclical data, and $0.37$ against $0.49$ and $0.52$ against $0.71$ on the cyclical data. ACTIVA is the weakest method on the pooled joint of both datasets, and the non-causal GMR baseline trails ours throughout. The margin is widest on the Variogram, the score most sensitive to cross-variable and cross-time dependence.

Where the conditioning baseline fails \emph{qualitatively} is the decomposition by causal role. Non-descendants are causally shielded from the intervention by definition, so their correct interventional target is their observational \emph{marginal} $p_\mathcal{M}(V_i)$, a distribution the model has effectively seen in context, making them an intrinsically easier target than descendants; a correct causal model should therefore improve from desc to nondesc. The informative quantity is the \emph{proportional} reduction. Our model reduces its Energy excess by $51\%$ from descendants to non-descendants on the non-cyclical data ($0.16$ to $0.08$), and by $56\%$ on the cyclical data. ACTIVA and GMR never achieve this: their excess falls by at most $9\%$, and on three of the four dataset-method pairs it is larger on names the shock cannot reach than on descendants. The MACE-TNP contrast isolates the mechanism: sharing the amortized causal framework and differing mainly in temporal structure, it exceeds our reduction on the non-cyclical data ($64\%$ against $51\%$) but retains only under half of it once the dynamics vary ($28\%$ against $56\%$). Selectivity is therefore reproducible across amortized causal estimators rather than an artifact of our decoder, but retaining it once the interventional law varies over the horizon requires the temporal structure. For GMR this is by construction, since it predicts the observational \emph{conditional} $p_\mathcal{M}(V_i \mid V_j)$ rather than the marginal and so assigns phantom effects to causally shielded names. We call this property \emph{causal selectivity}. 

Practically, this excess is the spurious response a stress test inherits. On shielded names the correct answer is the observational marginal, so excess over the Oracle floor there is not intrinsic difficulty but risk attributed to exposures the shock cannot reach. On the cyclical data, for example, GMR carries $0.59$ Energy of such excess against $0.07$ for ours.

\subsection{Horizon-resolved Predictive Quality}
\begin{figure}
    \centering
    \includegraphics[width=\linewidth]{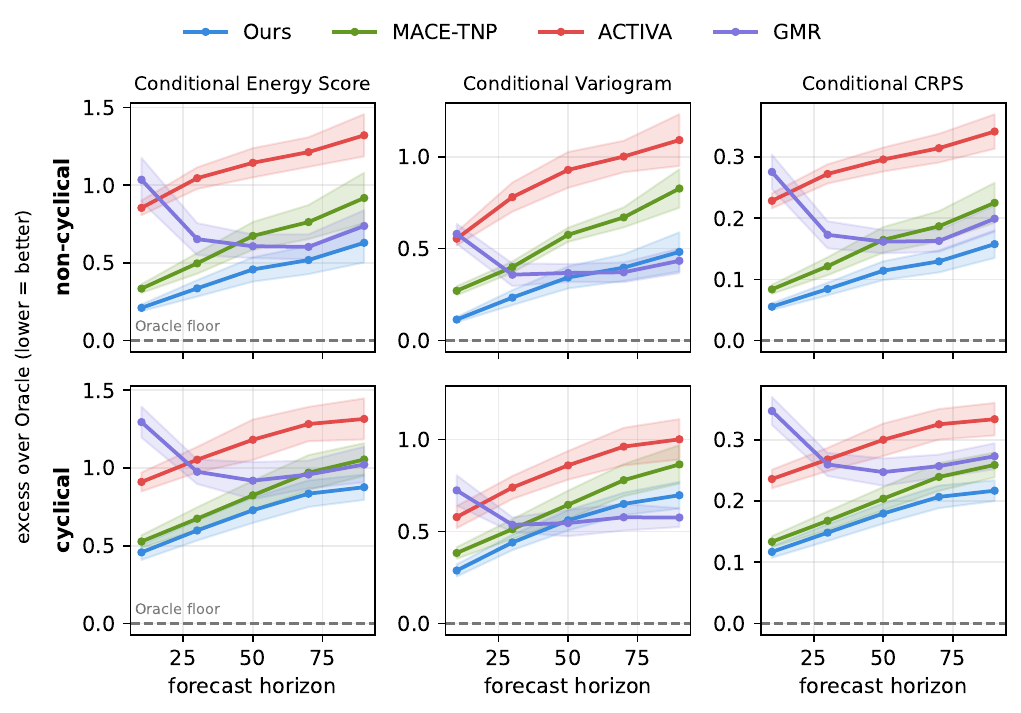}
    \caption[]{Horizon-resolved skill. Exact-time conditional Energy Score (left), Variogram Score (middle) and per-variable CRPS (right) versus forecast horizon, for the non-cyclical (top) and cyclical-mean (bottom) CIR datasets. Curves are \emph{excess over the Oracle floor} (dashed line at $0$); lower is better; bands are $95\%$ confidence intervals.}
    \Description{Grid of six line plots, two rows by three columns. Rows are
    the non-cyclical dataset on top and the cyclical dataset below; columns
    are conditional Energy Score, conditional Variogram and conditional CRPS.
    Each panel plots score excess over the Oracle floor, marked by a dashed
    line at zero, against forecast horizon from about 10 to 90, with four
    curves and shaded confidence bands. Our method is the lowest curve at
    every horizon in five of the six panels. ACTIVA is the highest curve
    throughout and rises steadily with horizon. MACE-TNP rises in parallel
    above ours. GMR behaves differently from the rest: it starts as the worst
    or near-worst method at the shortest horizon, falls steeply out to about
    horizon 30, then flattens and converges toward the other methods, dropping
    slightly below our curve at the longest horizons in the cyclical Variogram
    panel. The vertical separation between our curve and every baseline is
    widest at the shortest horizons and narrows as the horizon grows.}
    \label{fig:horizon_scores}
\end{figure}

Having established the whole-horizon joint picture, we now resolve performance by forecast horizon to probe temporal calibration. Figure~\ref{fig:horizon_scores} reports the exact-time conditional scores of Section~\ref{sec:metrics}, resolved by horizon, on both datasets, again as \emph{excess over the Oracle floor}. 
This exposes how each method's distributional fidelity evolves rather than collapsing temporal behaviour into a single number.

Reading the curves across all three scores reveals where the methods separate. At short horizons our method has the smallest Oracle gap on both datasets: on the cyclical data its near-horizon Energy excess is $0.46$, below MACE-TNP, ACTIVA, and GMR, and on the non-cyclical data it is $0.21$ against $0.33$ for MACE-TNP, with GMR and ACTIVA more than four times higher. GMR is weakest here, relative to its own longer-horizon predictions, because the initial phase of the intervened CIR is where the mean transitions from the observational to the intervened value, a shift it cannot capture by design since it is fitted to stable observational trajectories.

As the horizon grows the field converges: in the longest non-cyclical bin GMR becomes indistinguishable from our model within the plotted intervals, while ACTIVA remains far above throughout. Our model leads MACE-TNP in every bin on all three scores on both datasets, with the Energy and Variogram margins separated at the plotted intervals in every non-cyclical bin and in the near-horizon Variogram of the cyclical data, and stays ahead of GMR on Energy and CRPS in every bin of both datasets, up to the long-horizon Variogram, where GMR falls below us on both.

Taken together, the two axes separate the methods cleanly. The horizon-resolved view places our largest relative gains where temporal calibration is decisive, near the observed context, while the lead over MACE-TNP holds in every bin on both datasets, with the margin widest in the final bin. No baseline is strong on both axes across both datasets: GMR fails the causal selectivity of the joint, ACTIVA fails temporal calibration, and MACE-TNP, trails on both axes and retains under half its causal selectivity once the dynamics are time-varying. GMR's long-horizon strength on the conditional scores does not contradict this, since those scores ignore both cross-horizon mass allocation and causal selectivity. CIR-ACTIVA leads on both axes and both datasets.

\subsection{Ablation Study}
We validate our two core temporal design choices via an ablation (Table~\ref{tab:ablation}):
the \emph{expressive latent} (per-variable $L{=}8$ code with flow prior; $-$expr reverts to
$L{=}1$ with a diagonal GMM prior) and the \emph{bucketed decoder} (horizon-partitioned
mixture; $-$bucket replaces it with a plain joint GMM over $(\boldsymbol{Y},t)$, conditioned
analytically at the query time); \emph{base} compounds both.

Firstly, we observe that the two components together drive a clear near-horizon gain on both datasets: against \emph{base} the full model lowers the near-horizon Variogram from $0.341$ to $0.218$ on non-cyclical data and from $0.445$ to $0.384$ on cyclical data, with the non-cyclical near-horizon Energy and Variogram contrasts separated at the reported intervals. Second, $-$expr is consistently more costly than $-$bucket on both datasets. On the non-cyclical data removing both components costs more than removing either alone on every metric but the joint CRPS. Removing the expressive latent hurts the joint and near-horizon scores alike; removing the bucketed decoder hurts only at the near horizon, leaving the joint indistinguishable and its CRPS $0.002$ lower. On the cyclical data the three ablated variants are mutually unresolved at the reported intervals, so only their common gap to the full model is informative there. The full model is best or tied on eleven of the twelve entries and is thus the right operating point when horizon-specific calibration is the goal.

\begin{table}[t]
\centering
\small
\setlength{\tabcolsep}{4pt}
\caption{Ablation Results. Joint and near-horizon ($h\!\in\![0,20)$, exact-$t$ conditional) Energy (ES), Variogram (VS), and CRPS; subscripts are $95\%$ CI half-widths; lower is better. }
\label{tab:ablation}
\begin{tabular}{@{}l cccccc@{}}
\toprule
&\multicolumn{3}{@{}c}{\textit{Joint}}&\multicolumn{3}{@{}c}{\textit{Near}}\\
 & ES & VS &  CRPS &  ES &  VS &  CRPS \\
\midrule
\multicolumn{7}{@{}l}{ \textit{Non-cyclical data}} \\
Full       & $\mathbf{1.325}_{\pm.06}$ & $\mathbf{1.170}_{\pm.09}$ & $0.269$ & $\mathbf{0.588}_{\pm.02}$ & $\mathbf{0.218}_{\pm.01}$ & $\mathbf{0.156}$ \\
$-$expr.   & $1.418_{\pm.06}$ & $1.288_{\pm.08}$ & $0.300$ & $0.694_{\pm.02}$ & $0.288_{\pm.01}$ & $0.184$ \\
$-$bucket  & $1.325_{\pm.07}$ & $1.208_{\pm.09}$ & $\mathbf{0.267}$ & $0.660_{\pm.02}$ & $0.284_{\pm.02}$ & $0.172$ \\
base       & $1.419_{\pm.07}$ & $1.307_{\pm.09}$ & $0.298$ & $0.772_{\pm.03}$ & $0.341_{\pm.03}$ & $0.203$ \\
\addlinespace
\multicolumn{7}{@{}l}{ \textit{Cyclical data}} \\
Full       & $\mathbf{1.503}_{\pm.06}$ & $\mathbf{1.419}_{\pm.06}$ & $\mathbf{0.320}$ & $\mathbf{0.822}_{\pm.05}$ & $\mathbf{0.384}_{\pm.03}$ & $\mathbf{0.212}$ \\
$-$expr.   & $1.532_{\pm.05}$ & $1.473_{\pm.05}$ & $0.328$ & $0.899_{\pm.05}$ & $0.455_{\pm.03}$ & $0.231$ \\
$-$bucket  & $1.525_{\pm.06}$ & $1.449_{\pm.07}$ & $0.325$ & $0.872_{\pm.05}$ & $0.438_{\pm.03}$ & $0.223$ \\
base       & $1.525_{\pm.06}$ & $1.442_{\pm.05}$ & $0.327$ & $0.894_{\pm.05}$ & $0.445_{\pm.03}$ & $0.230$ \\
\bottomrule
\end{tabular}
\end{table}

\section{Related Work}
Time-series causal discovery spans Granger-style predictability tests~\citep{Granger1969InvestigatingMethods}, non-Gaussian identifiability~\citep{Hyvarinen2010EstimationNon-Gaussianity,Peters2013CausalModels}, constraint-based discovery under autocorrelation and non-stationarity~\citep{Runge2019DetectingDatasets,Huang2020CausalData}, amortized graph recovery and foundation-model pretraining for discovery~\citep{Lowe2022AmortizedData,Stein2024EmbracingData}. In all cases the output is a graph or disentangled representation, not the interventional distribution of future trajectories. Closer to our goal, recent amortized methods shift from graph recovery to direct effect estimation: \citet{Lorch2022AmortizedLearning} amortize causal structure inference via variational inference over graphs, while CausalPFN~\citep{Balazadeh2025CausalPFN:Learning}, our base model ACTIVA~\citep{Sauter2025ACTIVA:Autoencoder}, and our baseline MACE-TNP~\citep{Dhir2026EstimatingMeta-Learning} amortize causal effect estimation itself, via in-context learning over a prior of SCMs, a $\beta$-CVAE, and the Transformer Neural Process framework~\cite{Nguyen2022TransformerModeling}, respectively. None builds in a time-series-specific inductive bias, the gap our time-indexed distributional forecasting fills: CIR-ACTIVA extends the ACTIVA $\beta$-CVAE with temporal structure, while CausalPFN, moreover, targets non-temporal tabular effects.

A distinct literature targets causal effect estimation from longitudinal observational data, either by learning balanced temporal representations for counterfactual outcome prediction under time-varying treatment~\cite{BicaEstimatingRepresentations,melnychuk2022causal}, or by embedding causal semantics into diffusion or VAE generators for interventional density modeling and data synthesis~\citep{Lorch2024CausalDiffusions,Xia2025CausalModels,Thumm2025TowardsSimulators}. Across this strand, methods either require the causal graph at test time, retrain per dataset, or produce synthetic data rather than amortized distributional effect estimates.

Probabilistic generative forecasters like flow matching~\citep{El-Gazzar2025ProbabilisticMatching} and coupled diffusion-VAE models~\citep{LiGenerativeDisentanglement} learn conditional future trajectory distributions but answer predictive, not interventional, queries and retrain per dataset. Foundation-style models push zero-shot predictive performance via large-scale pretraining~\citep{Hoo2025TheFeatures,DooleyForecastPFN:Forecasting}; none answer interventional queries or jointly model time as a random variable alongside the target series.

Within the financial domain, credit-spread and CDS forecasting spans Markov-switching autoregressions and SVMs~\cite{Vukovic2022AreAutoregression}, Merton-model-augmented LSTMs~\cite{Mao2023ForecastingLSTMs}, MacroVAE~\cite{Kubiak2025MacroVAE:Conditioning}, and attention-based recurrent networks~\cite{Liu2022AMechanism}. All estimate the historical joint $p(\boldsymbol{V}\mid\text{past})$, not the interventional distribution $p(\boldsymbol{V}\mid do(V_j=x))$ demanded by causal forecasting.

Across all these strands, no prior work combines amortized meta-learning over a distribution of SCMs with time-series-specific inductive bias and distributional interventional forecasting for multivariate financial time series; this is the gap our approach fills.


\section{Conclusion}
We presented CIR-ACTIVA, a model for amortized interventional forecasting in multivariate CIR systems, together with a causal multivariate CIR process that supplies the interventional ground truth real markets cannot. Given observed  trajectories and an intervention that stresses one series, it forecasts calibrated distributions of the system's causal response at multiple horizons. It leads every baseline in our evaluation on both causal selectivity and horizon-resolved calibration, with the largest gains at the short horizons where the causal response is most pronounced.

Relying on synthetic ground truth is both a strength and a limitation. It is a strength because our method's ability to recover the data-generating interventional distribution, is established on the simulated samples and holds regardless of how closely the simulator resembles any real market. Because the model itself consumes only CIR-type trajectories, agnostic to what the calibrated parameters represent, that validity extends beyond CDS to the other mean-reverting series CIR describes, such as short rates, default intensities, and stochastic volatility, and to interventional tasks beyond stress testing. The limitation is that synthetic ground truth alone says little about real spread behavior. Backtesting narrows this gap: our simulated traces match real spread levels but understate shock persistence and the frequency of large moves. Both deficits make the synthetic task the easier one, so all our scores bound real-data performance from above. 
The experiments cover systems of three names; scaling to portfolio-sized panels, where self-attention over variables becomes expensive, is a separate question.

More broadly, our results show that observational forecasting accuracy is not a proxy for interventional correctness in coupled financial systems, where the error takes the concrete form of phantom shock responses on names the shock cannot reach. CIR-ACTIVA addresses this by amortizing over a distribution of causal CIR tasks, delivering interventional predictions without retraining for each scenario. Beyond the model, the causal CIR simulator contributes a controlled benchmark for distributional causal effect estimation against which future causal forecasters can be measured.

\ifarxiv
\begin{acks}
This research was partially funded by the Hybrid Intelligence Center, a
10-year programme funded by the Dutch Ministry of Education, Culture and
Science through the Netherlands Organisation for Scientific Research,
\url{https://hybrid-intelligence-centre.nl}, grant number 024.004.022.
\end{acks}
\fi

\bibliographystyle{ACM-Reference-Format}
\bibliography{references}

\end{document}